\RequirePackage{fix-cm}
\documentclass[smallextended]{svjour3}       
\smartqed  
\usepackage{graphicx}
\usepackage[T1]{fontenc}
\usepackage[utf8]{inputenc}
\usepackage{multirow}

\begin{document}

\title{Comparative Analysis of Predictive Methods for Early Assessment of Compliance with Continuous Positive Airway Pressure Therapy}

\titlerunning{Comparative Analysis of Predictive Methods...}        

\author{Xavier Rafael-Palou \and Cecilia Turino \and Alexander Steblin \and Manuel S\'{a}nchez-de-la-Torre \and Ferran Barb\'{e} \and Eloisa Vargiu
}


\institute{Xavier Rafael-Palou, Alexander Steblin, Eloisa Vargiu \at
              Eurecat Technology Center - eHealth Unit - Barcelona (Spain)\\
              \email{\{xavier.rafael, alexander.steblin, eloisa.vargiu\}@eurecat.org} 
           \and
           Cecilia Turino, Manuel S\'{a}nchez-de-la-Torre, Ferran Barb\'{e} \at
           Institut de Recerca Biomèdica (IRBlleida), Lleida and CIBERES - Madrid (Spain) \\
              \email{\{bloom6784, sanchezdelatorre\}@gmail.com, febarbe.lleida.ics@gencat.cat}
}

\date{Received: date / Accepted: date}

\maketitle

\begin{abstract}
Patients suffering obstructive sleep apnea are mainly treated with continuous positive airway pressure (CPAP). A good compliance with this therapy is broadly accepted as \(>\)4h of CPAP average use nightly. Although it is a highly effective treatment, compliance with this therapy is problematic to achieve with serious consequences for the patients’ health. Previous works already reported factors significantly related to compliance with the therapy. However, further research is still required to support clinicians to early anticipate patients’ therapy compliance. This work intends to take a further step in this direction by building compliance classifiers with CPAP therapy at three different moments of the patient follow-up (i.e. before the therapy starts and at months 1 and 3 after the baseline). Results of the clinical trial confirmed that month 3 was the time-point with the most accurate classifier reaching an f1-score of 87\% and 84\% in cross-validation and test. At month 1, performances were almost as high as in month 3 with 82\% and 84\% of f1-score. At baseline, where no information of patients$’$ CPAP use was given yet, the best classifier achieved 73\% and 76\% of f1-score in cross-validation and test set respectively. Subsequent analyzes carried out with the best classifiers of each time point revealed that certain baseline factors (i.e. headaches, psychological symptoms, arterial hypertension and EuroQol visual analog scale) were closely related to the prediction of compliance independently of the time-point. In addition, among the variables taken only during the follow-up of the patients, Epworth and the average nighttime hours were the most important to predict compliance with CPAP.

ClinicalTrials.gov NCT03116958. April 2017, retrospectively registered.

\end{abstract}

\section{Background}\label{sec:background}
Obstructive sleep apnea (OSA) \cite{bib:smith94} is defined as repeated episodes of shallow or paused breathing during sleep, despite the effort to breathe. This syndrome is caused by complete or partial obstructions of the upper airway leading to daytime sleepiness and impaired cardiopulmonary function. Gold standard treatment of OSA involves the use of a device that administers continuous positive pressure (CPAP) in the respiratory tract of patients \cite{bib:kribbs93}. Continuous positive pressure is applied to the upper airway with a nasal mask, nasal prongs, or a mask that covers both the nose and mouth \cite{bib:prosise94} \cite{bib:sanders94}. A pneumatic splint is provided that prevents narrowing and closure of the upper airway regardless of the site of obstruction \cite{bib:sullivan81}. The level of positive pressure required to sustain patency of the upper airway during sleep should be determined in the sleep laboratory.

Several studies aimed at objectively assess the effects of using CPAP show that this kind of intervention is highly effective in improving symptoms, such as daytime sleepiness, morbidity, and mortality rates related to cardiovascular diseases \cite{bib:Engleman341} \cite{bib:marin2005long}. Although it is highly effective in minimizing OSA symptoms, up to 36\% of patients do not use or even discontinue CPAP \cite{bib:hussain2014compliance} \cite{bib:campos2016long}.


Adherence to therapy is thought to be influenced by a complex array of, as yet, poorly characterized factors. Patient characteristics such as age, sex, and marital status have not consistently been shown to be predictors of CPAP adherence \cite{bib:weaver2007continuous}. Severity of OSA, as determined by the apnea-hypopnea index (AHI), oxygen desaturation index (ODI), or daytime sleepiness, have been shown to have a weak relationship with CPAP use in some studies \cite{bib:mcardle1999long}\cite{bib:kohler2010predictors}\cite{bib:krieger1996long}. Furthermore, the adherence measured  3 days after CPAP initiation seems to be a good predictor of long-term adherence \cite{bib:budhiraja07}.

Despite these efforts, the reasons that lead to compliance with therapy remain an open field of research. Therefore a prompt detection of adherence pattern would reduce its misuse and abandonment ratios and allow a specific approach to improving compliance. Unfortunately, there is a clear lack of clinical analytical tools to support the early prediction of compliant patients.



When statistically analyzing possible factors that might determine CPAP compliance, there are common complexities that compromise predictive capacity and robustness of  finding. The limited number of participants, the large number of clinical variables and the quality of collected data are just to name a few. To overcome these limitations the use of machine learning (ML) might be an alternative solution to the aforementioned problems. Despite  the numerous ML applications in the medical domain, (e.g. disease diagnosis \cite{bib:ccinar2009early}\cite{bib:hassanien2014mri}), compliance with therapy is usually constrained to the medication adherence  problem \cite{bib:lee2013predictors}\cite{bib:bourdes2011prediction}. Recent ML algorithms (e.g. support vector machines \cite{bib:cortes95} and artificial neural networks \cite{bib:bishop1995neural}) often provide highly accurate predictive models. However, such models lack transparency and therefore their interpretation is difficult \cite{bib:auria2008support}. As a consequence, other classification algorithms in the medical field are still preferred (e.g. logistic regressions \cite{bib:cox58} or decision trees \cite{bib:rokach2014data}).

In summary, the present study has the following objectives. On one hand, to provide a comparative analysis of predictive methods of CPAP compliance built using machine learning techniques in different stages of treatment. On the other hand, to define the most important factors associated with CPAP compliance identified by the best predictive methods obtained in the different initial stages of therapy.

\section{Methods}\label{sec:methods}
\subsection{Participants}

51 adult patients ($>$18years), diagnosed with OSA (15 or more apneas/hypopneas per hour in an overnight sleep study) and requiring CPAP treatment were recruited at Hospital Arnau de Vilanova (Lleida, Spain). Patients with impaired lung function (overlap syndrome, obesity hypoventilation, and restrictive disorders), severe heart failure, psychiatric disorders, periodic leg movements, pregnancy, other dyssomnias or parasomnias, and/or a history of previous CPAP treatment were excluded. The study was approved by the hospital ethics committee (Approval number: CEIC-1283). All recruited patients signed an informed consent form.

Of the 51  patients originally included in the study, 3  were excluded due to malfunction of the CPAP machine, 5 did not attend the last visit at the sleep unit and 1 patient died during the study.

The final sample consisted of 42 patients (29 males and 13 females) with a mean age of 56.93+/-12.58 yrs. Their BMI was 33.83+/-6.46 and their number of apnea or hypopneas per hour of sleep (apnea/hypopnea index or AHI) was 53.13+/-20.72 events/h. In our sample, 60\% (25) of all patients were active workers and 33\% (14) were retired. The sample also had 62\% of nonsmokers (26) and 57\% of nonalcohol consumers (24). In terms of CPAP device use, the patients scored an average of nightly hours of use of 5.44+/-1.74 at month-1, 5.33+/-1.90 at month-3, 5.07+/-2.10 at month-6.  

\subsection{Datasets}
The study variables from the 42 patients were manually collected by lung specialists  along four visits at month-0 (baseline or T0), at month-1 (T1), at month-3 (T3) and at month-6 (T6).
During the first visit clinicians gathered 77 features organized in five categories: clinical history (e.g. depression, anxiety, arterial hypertension (HTA), cardiopathy, neurological disease, respiratory disease), symptoms (e.g. irritability, apathy, depression, insomnia), co-morbidities (e.g. diabetes, obesity, dyslipidemia), therapies (e.g. beta blockers, diuretics), sleeping test (e.g. sleeping time, AHI, percentage of the night spent with  oxygen saturation $<$ 90\% or CT 90) and basal information (e.g. size, weight, BMI, tas, tad, oxygen saturation). In the second visit, when the patients had the CPAP machine at home during one consecutive month, 16 new features related to monitoring were collected (e.g. nightly average use, abandon or adverse effects of the treatment, such as dry mouth, allergies, and cutaneous irritations). At the third month (T3), the same number of features as in T1 were gathered but adding 5 new ones (i.e. size, weight, BMI, drugs removed and drugs added). At month-6, although some other variables were collected, for the purpose of this study only the average use of nigh hours was considered. Eventually, three datasets (D0, D1, and D3) with an incremental number of features (i.e. D1 features = D0 features + features collected at T1) were created  with 77, 93 and  114 features. The full list of variables is described in table s1 of the supplementary material.

In this study, we addressed CPAP compliant users as those who had more than 4 hours on average per night during the first 6 months of treatment. Therefore, all samples from each dataset (ds) were labeled using the collected information about nightly hours/use on average of the CPAP device at the end of the month-6. In so doing, 24 (57\%) patients were labeled as 'compliant', class '1', as they correctly followed the CPAP therapy prescription (more than 4h nightly on average). On the contrary, 18 (43\%) patients did not achieve the prescribed treatment (minus or equal than 4 hours nightly on average) and they were labeled as 'non-compliant', class '0'. 

\subsection{Preprocessing}
Datasets D0, D1, and D3 collected at time points T0, T1 and T3, respectively, were statistically described for a better understanding of the sample. In this task, the Mann-Whitney U test was used to evaluate the statistical significance of quantitative variables with CPAP compliance and Chi-square tests for qualitative characteristics. Previously, the categorical features were converted into numerics to achieve a homogeneous data type sample. Variables with only two categories were directly mapped into binary values. Variables with more than two categories, given their underlying incremental meaning, were mapped into unsigned ordinal values.

Afterwards, we carried out a set of preprocessing tasks to reduce possible noise and redundancy in the datasets.  First, given the existence of null values in the datasets, an imputation process was carried out consisting of computing the mode for the categorical characteristics and the mean for the numerical characteristics. Subsequently, the distributions of the categorical characteristics were analyzed, which revealed variables with few individuals by category.  The features with a number of individuals below a threshold were removed from the study to avoid the noise they might introduce when building predictive models. To catch up possible information redundancy in the datasets, we computed the mutual information \cite{bib:ross2014mutual} score for the categorical variables in a pair-wise manner. From each pair, we kept the variable more statistically significant with the dependent variable using Chi-square test. Among the numerical features, we applied a correlation analysis to detect highly redundant features. Given the existence of non-normally distributed numerical features, we used the Spearman correlation method on all numerical variables in a pair-wise manner. Empirically we set-up a threshold for the correlation scores above of which one feature of the pair was removed (i.e. the feature with the highest p-value). P-values less than 0.05 were considered statistically significant.

\subsection{Classification Framework}
All preprocessed datasets presented common particularities such as a small number of samples, the presence of missing values, class unbalance and high multidimensionality feature space. To cope with these complexities we designed a classification framework flexible enough to enable the execution of heterogeneous pipelines or sequence of configurable machine learning steps. In particular, the pipelines were composed of three mandatory steps (i.e. imputation, variance filtering and data standardization), two optional steps (i.e. feature selection and feature sampling) and two more final steps (i.e. classifier training and evaluation). In total 80 pipelines were configured from 4 feature selection methods, 5 classifier algorithms, 2 sampling strategies and 2 evaluation metrics. 

\begin{figure}[h!]
  \includegraphics[width=1.0\textwidth]{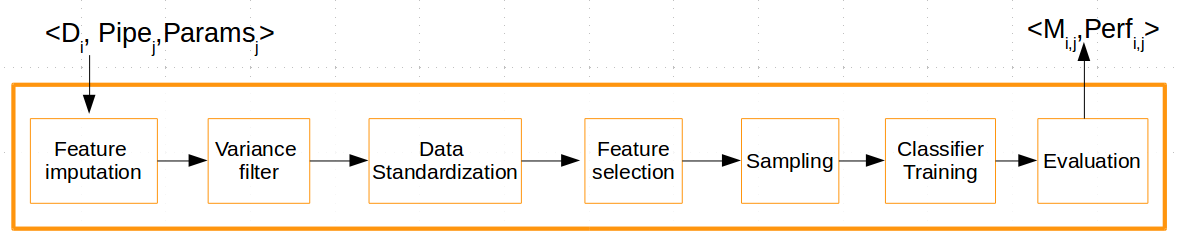}
  \caption{Pipeline steps designed for building classifiers for compliance with the CPAP therapy.}
  \label{fig:pipe}
  \end{figure}

The result of running (i.e. training or building) a pipeline (Pipe$_i$) on a dataset ($D_j$) with parameters ($params_i$) is a predictive model or classifier ($M_{i,j}$) with its associated predictive performance (Perf$_{i,j}$). Figure \ref{fig:pipe} shows the scheme with the inputs, outputs and the different steps that configure the pipelines for compliance with CPAP therapy. 

The first step of a pipeline is the imputation of null values. To do this, given the small proportion of null values in the datasets, a simple strategy was proposed to replace the null values with their most frequent value (for categorical characteristics) and with the mean value (for numerical characteristics).

The second step consists of a simple filter method to eliminate features with zero variance, that is, to eliminate these characteristics that have the same value in all the samples and that do not provide any additional information to the data set.

Since the data come from different sources, the next step is to standardize the data. This step consisted of homogenizing all features to zero mean and variance one. This transformation step is crucial for the construction of many classification algorithms since it allows them to compare features without harming their performance or execution time \cite{bib:inza2010machine}.

Feature selection (fs) was introduced in the pipelines given a large number of features compared with the number of samples for each dataset ($ p > n $). This type of methods aims to reduce over-fitting by improving model performance and generalization, to provide faster and more cost-effective models and simplify models making them easier to interpret \cite{bib:bermingham2015application}. Feature selection methods are usually divided into three categories: filter, wrapper, and embedded \cite{bib:guyon2003introduction}. Filter methods, in general, examine features individually with the class, wrapper methods use a predictive model to generate subsets of features evaluated according to their predictive power, and embedded methods search for an optimal subset of features during the training of the prediction model. In this study, we used one method for each of the different feature selection strategies. For the filter-based strategy, we defined a simple method (\textit{combine\_fs}) that makes a ranking of the features by their statistical significance with the class (i.e. applying ANOVA or chi-squared tests according to the data type of the characteristics). Then, this method returns the subset of features through a configurable threshold. For the wrapper strategy, we proposed the recursive feature elimination (\textit{rfe\_rf\_fs}) method \cite{bib:golub99} configured with a random forest to provide the importance of the features. The embedded strategy was entitled to the application of a linear model configured with the L1 norm (\textit{lasso\_fs}) \cite{bib:tibshirani1996regression}. It was also considered the possibility of not using any method of feature selection.

The next step is data sampling (sm). In particular, we proposed the use of the smote sampling method \cite{bib:chawla2002smote}. This method consists of creating synthetic samples (i.e. detecting similar instances and performing small perturbations in their values) of the under-represented class samples instead of creating copies, as the over-sampling method would do. The main idea behind this method is to avoid the bias produced by many standard classifier learning algorithms towards the class with a larger number of instances.

Regarding the training and evaluation stage, we selected several classification algorithms (cls) to deal with various classification strategies (i.e. linear, non-linear, distance-based and tree-based). In fact, the provision of different classification strategies is especially appropriate in complex datasets when the distribution of data is not clearly manifested. As early mentioned in the paper, the interpretability of the resulting predictive models is also a desired condition. Therefore, we opted for logistic regression (LR) \cite{bib:cox58}, k-nearest neighbor ($ k-NN $) \cite{bib:altman92} and random forest ($RF$) \cite{bib:ho95} for the subset of interpretable classification algorithms (termed descriptive within the study). In contrast, we chose support vector machines ($ SVM $) \cite{bib:cortes95} and artificial neural networks ($NN$) \cite{bib:bishop1995neural} for the subset of algorithms with less interpretative capacity but with a potential greater discriminatory capacity (i.e. referred to as non descriptive).

\subsection{Evaluation Setup}
In order to ensure adequate performance evaluation, the available data were stratified and randomly divided into train (29 rows, 12 non-compliant and 17 compliant) and test (13 rows, 6 non-compliant and 7 compliant) sets with a ratio of 70/30. Therefore, the training set partitions of the three data sets (D0, D1, and D3) contained the same individuals. The same rule applies for the test set.

Test sets remained untouched until the end of the process. Training sets were used for 10-fold cross-validation to enable proper model tuning and evaluation. This technique is particularly suitable when the sample size is small. Indeed, as suggested in \cite{bib:friedman2001elements} the entire sequence of processes that composed each pipeline was wrapped-up within the cross-validation technique in order to reduce the possibility of obtaining too optimistic or pessimistic results. Thus, training data were randomly split into stratified train-validation sets (20 rows, 8 non-compliant and 12 compliant) and stratified test-validation sets (9 rows, 4 non-compliant and 5 compliant) following a ratio of 70/30. Then, for each of the configured pipelines (i.e. 80 pipelines), we created as many experiments as combinations of values for the different hyper-parameters defined for each method of the pipeline (table \ref{tab:params}). 

\begin{table}[ht!]
\caption{Different pipeline parameters tested using grid-search and 10-fold CV.}
\label{tab:params}
\begin{center}
\begin{tabular}{r@{\quad}ll}
\hline
\multicolumn{1}{c}{\rule{0pt}{12pt}Pipeline Step}&\multicolumn{2}{c}{Parameter Options}\\[2pt]
\hline\hline\rule{0pt}{12pt}
Combine\_fs & percentile = [5, 10, 20, 30, 40, 50]& \\  \hline
Lasso\_fs 
& \begin{tabular}{l}
	estimator = Logistic Regression  \\
	penalty = "l1" \\
	C = [5, 10, 20, 30, 40, 50]
\end{tabular} & \\ \hline
RFE\_RF\_fs      
& \begin{tabular}{l}
	class\_weight = 'balanced' \\
    n\_estimators = {100} \\
    step = {[}0,1 {]} \\
    n\_features\_to\_select = {[}0.4,0.6,0.8{]}
\end{tabular} & \\ \hline
Smote\_fs 
& \begin{tabular}{l}
	n\_neighbors = [3,4,5] \\
	ratio='auto' \\
	kind='regular'
\end{tabular} & \\ \hline
K-NN         
& \begin{tabular}{l}
	n\_neighbors = [1,3,5,7,9,11] \\
    weights = ['uniform', 'distance']
\end{tabular} & \\ \hline
LR 
& \begin{tabular}{l}
C = [0.00001,0.0001,0.0005,0.001,0.005,0.01,0.05,0.1,0.5,1,5,10,15,30] \\
class\_weight = [None, 'balanced'] \\
penalty = ['l1', 'l2']
\end{tabular} & \\ \hline
RF          
& \begin{tabular}{l}
  n\_estimators = [100,150,200,250,500] \\
  criterion = ['entropy','gini'] \\
  max\_depth = ['None',4,6] \\
  class\_weight = [None, 'balanced']
\end{tabular} & \\ \hline
SVM  
& \begin{tabular}{l}
  C = [0.01,0.1,0.5,1,5,10,15,30,50] \\
  gamma = [0.0001,0.001,0.01, 0.1,1,5] \\
  kernel = 'radial' \\
  class\_weight = [None, 'balanced']
\end{tabular} & \\ \hline
NN  
& \begin{tabular}{l}
alpha = [$1e-5$,0.00001,0.0001,0.001,0.01,0.1,1,3,5,10] \\
hidden\_layer\_sizes = [(30,),(50,),(70,),(100,),(150,), \\
(30,30),(50,50),(70,70),(100,100), \\
(30,30,30),(50,50,50),(70,70,70)]
\end{tabular} & \\ \\[2pt]
\hline
\end{tabular}
\end{center}
\end{table}

We performed 10-fold cross-validation for all experiments in each pipeline. This process was repeated twice, one for each of the proposed learning metrics (i.e. f1-weighted and precision-weighted). The learning metric, f1-weighted ($f1$), was selected since it is a suitable measure for unbalanced datasets. This metric combines the precision and recall metrics weighted by the number of samples per each class. The other selected metric, precision-weighted ($prec$), tends to prefer classifiers with less incorrect compliant predicted patients, which indeed they are the most harmful cases to avoid. This metric computes the ratio of correctly classified cases (i.e. compliant patients) among all positive classified cases weighted by the number of samples per each class. 

As a result of this 10-fold cross-validation process, we reported for each experiment the average and standard deviation performance of the learning metric with which the pipeline was configured. A greedy-search strategy was applied to select the best experiment, i.e. best pipeline parameterization. Additionally, with the intention to avoid the possible bias introduced in this process \cite{bib:cawley2010over}, we evaluated each best pipeline parameterization (i.e. pipelines with the appropriated values for their hyper-parameters) using a final outer stratified 10-fold cross-validation on the training data (i.e. with learning metric=f1-weighted, ratio=70/30 for cv-train and cv-test). As a result of this process we reported the final cross-validation performance (f1-score) of the pipelines. This score was provided by the f1-weighted metric since although having a high precision is desirable, a high recall (rec) is also needed especially for health institutions since it reduces false negatives and thus non-necessary clinical interventions and additional costs. 

This whole process was repeated 80 times for all pipelines of each dataset. The best pipelines of each dataset were identified by ranking the cross-validation performances (i.e. f1-score) reported by each pipeline. The best pipelines of each dataset were compared together in order to find statistically significant differences. We did the same among the best descriptive and non-descriptive pipelines. To do this, we used a 10-fold cross-validated paired t-test.

To complete the reporting of this analysis, we computed on the test set and for each of the best pipelines of each dataset a comprehensive set of scores (i.e. f1, precision, recall, AUC and confusion matrix) to enable a better understanding of the results. Also, we reported their ROC and learning curves.

\subsection{Feature Importances}
We reported the most important features from the best descriptive pipelines (in this study we only used those configured with random forest and logistic regression) for each dataset. To do this, we performed a ranking of features using a stability score \cite{bib:meinshausen2010stability}. This score measures how "stable" are the features of a predictive model. For this, we build $n$ times a pipeline using $n$ random subsets of fixed size $s$.

To compute this score, we created ($n=100$) randomly stratified partitions from the ($s=70\%$) of the entire dataset. For each data partition, we trained the best pipelines and keep a record of the selected features of the classifier and their weights (or feature importances for RF classifier). With this information, we computed the number of times any feature was selected (i.e. stability score) and its normalized absolute average weight (or importance). Since one classifier might report all features as relevant (i.e. non-zero), a threshold ($t > 0.4$) in the weights was empirically defined to make usable the stability score.

\section{Results}\label{sec:results}
\subsection{Data preprocessing}

A descriptive analysis of the initial data sets was carried out and summarized in tables (s2,s4,s6) of the supplementary material. In total (11/27/42) features had null values with ratios between 2.3\% and 12\% from the total number of rows. After the null imputation, we found (14/7/10) variables with underrepresented categories ($ <= 10\%$ of rows per category ). We also detected 4 pairs of categorical features (i.e. no active, anti-depressives, ADO and memory disorders) in the D0 with MI scores above 50\%. From the correlation analysis applied on the numerical variables, we found 4 highly ($ > 80\% $) redundant features in D0 (i.e. abdomen and hip circumference, weight and  CT90\%) and 4 features in D3 (i.e. size\_3, weight\_3, bmi\_3 and total\_use\_hours\_3). After the removal of these features, the final datasets were composed of (54/63/70) variables. 

\subsection{Classification analysis}
Eventually, we evaluated 76 out of 80 initially configured pipelines for each dataset. In particular, we rule out pipelines which had same classification algorithm (i.e. random forest) for feature selection and classification given their initial poor contribution to the experimental results and the long runtime required to complete their evaluation. 

Best pipelines (p0, p1, p3) for D0, D1 and D3 achieved 0.73+/-0.18, 0.82+/-0.06, 0.87+/- 0.15 of f1-score in cross-validation and 0.76, 0.84, 0.84 in test set (Table \ref{tab:topCls}). These pipelines were configured with precision-weighted metric and SVM algorithm for the D0 dataset; with smote sampling, f1-weighted metric and an SVM for the D1 dataset and with lasso feature selection, smote sampling, precision-weighted metric and an RF for the D3 dataset. 

\begin{table}[ht!]
\caption{Performances of the best pipelines in each dataset.}
\label{tab:topCls}
\begin{center}
\begin{tabular}{lllllll}
\hline
id & ds &     sm &        fs &              metric &     cls &                         params \\
\hline
p0 & D0 &   none &      none &  precision\_weighted &  SVM &          [0.001, balanced, 30] \\
p1 & D1 &  Smote &      none &         f1\_weighted &  SVM &           [0.001, None, 4, 15] \\
p3 & D3 &  Smote &  Lasso\_fs &  precision\_weighted &  RF &  [1, 250, gini, 4, None, None] \\
\\
\end{tabular}

\begin{tabular}{llllllll}
\hline
id &      cv\_prec &       cv\_rec &        cv\_f1 & test\_prec & test\_rec  & test\_f1 \\
\hline
p0  &   0.78+/-0.2 &  0.74+/-0.17 &  0.73+/-0.18 &      0.77 &     0.85 &    0.76 \\
p1 &  0.84+/-0.06 &  0.82+/-0.05 &  0.82+/-0.06 &      0.85 &     0.88 &     0.84 \\
p3 &  0.89+/-0.14 &  0.88+/-0.14 &  0.87+/-0.15 &      0.85 &     0.88 &     0.84 \\
\hline
\end{tabular}
\end{center}
\end{table}

\begin{figure}[h!]
  \includegraphics[width=1.0\textwidth]{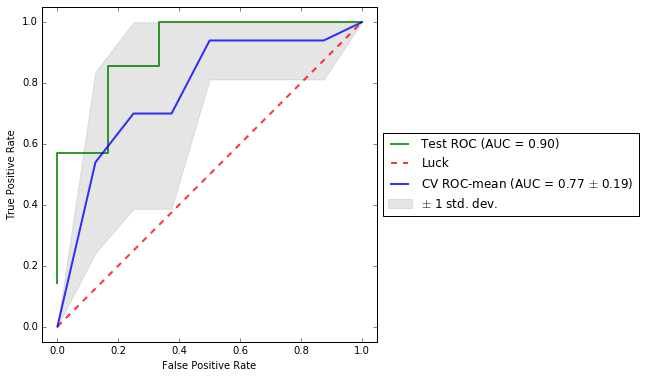}
  \caption{ROC curves for cross-validation and test of the best pipeline for dataset D0.}
  \label{fig:d0Roc}
      \end{figure}

\begin{figure}[h!]
   \includegraphics[width=1.0\textwidth]{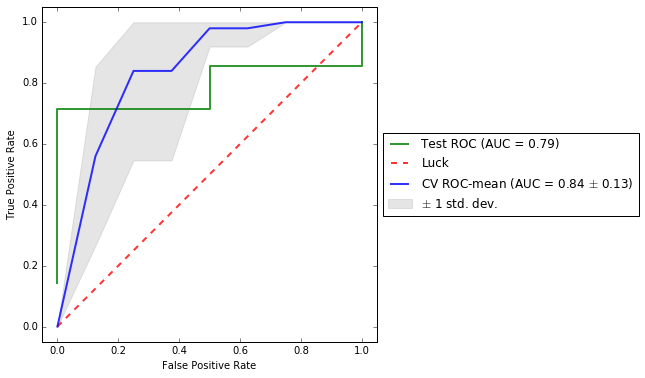}
  \caption{ROC curves for cross-validation and test of the best pipeline for dataset D1.}
  \label{fig:d1Roc}
      \end{figure}

\begin{figure}[h!]
   \includegraphics[width=1.0\textwidth]{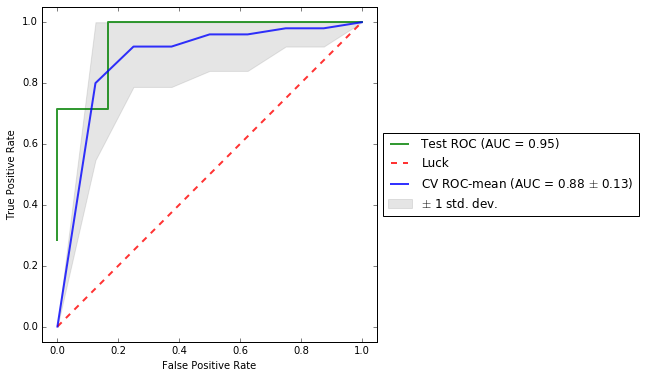}
  \caption{ROC curves for cross-validation and test of the best pipeline for dataset D3.}
  \label{fig:d3Roc}
      \end{figure}

\begin{figure}[h!]
	\includegraphics[width=1.0\textwidth]{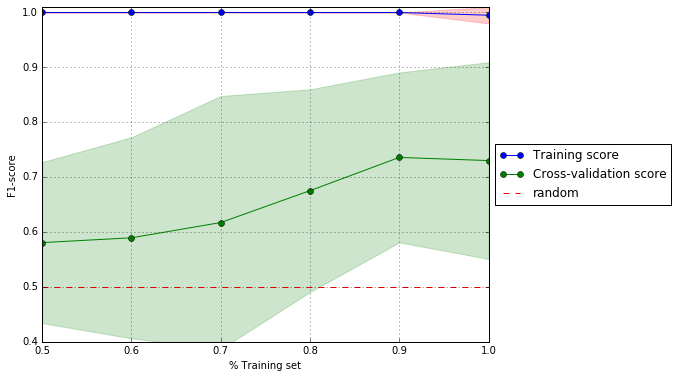}
  \caption{Learning curves of the best pipeline for dataset D0.}
  \label{fig:d0Learn}
      \end{figure}

\begin{figure}[h!]
   \includegraphics[width=1.0\textwidth]{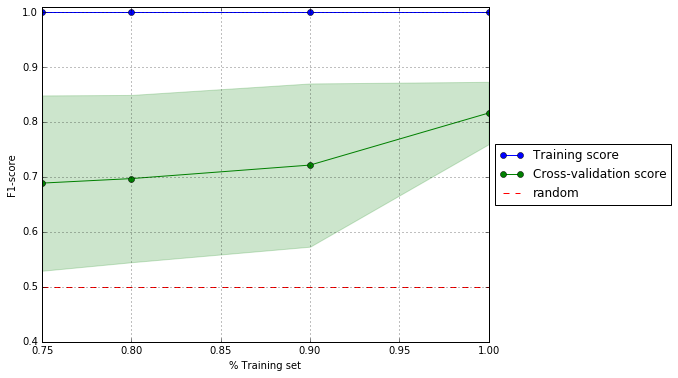}
  \caption{Learning curves of the best pipeline for dataset D1.}
  \label{fig:d1Learn}
      \end{figure}

\begin{figure}[h!]
      \includegraphics[width=1.0\textwidth]{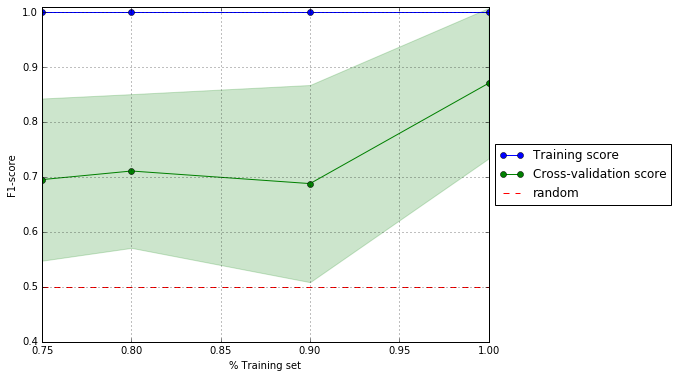}
  \caption{Learning curves of the best pipeline for dataset D3.}
  \label{fig:d3Learn}
      \end{figure}

To visually support these values, figures (\ref{fig:d0Roc}, \ref{fig:d1Roc}, and \ref{fig:d3Roc}) show the area under the receiver operating characteristic (ROC) curves and figures (\ref{fig:d0Learn}, \ref{fig:d1Learn}, and \ref{fig:d3Learn}) show the learning curves with the effects of increasing the size of the training set in their performances. 

\begin{table}[ht!]
\caption{Performance comparison between best pipelines for each dataset.}
\label{tab:diffBestPipes}
\begin{center}
\begin{tabular}{lllllll}
\hline
Pipelines &     Difference (cv\_f1) &        statistic &              p\_value \\
\hline
p0 vs p1 &   0.09 +/- 0.15 &      -1.71 &  0.1201 \\
p0 vs p3 &   0.14 +/- 0.18 &      -2.70 &  0.0241 \\
p1 vs p3 &   0.05 +/- 0.14 &    -1.0931 &  0.3027
\end{tabular}
\end{center}
\end{table}

\begin{figure}[h!]
  \includegraphics[width=1.0\textwidth]{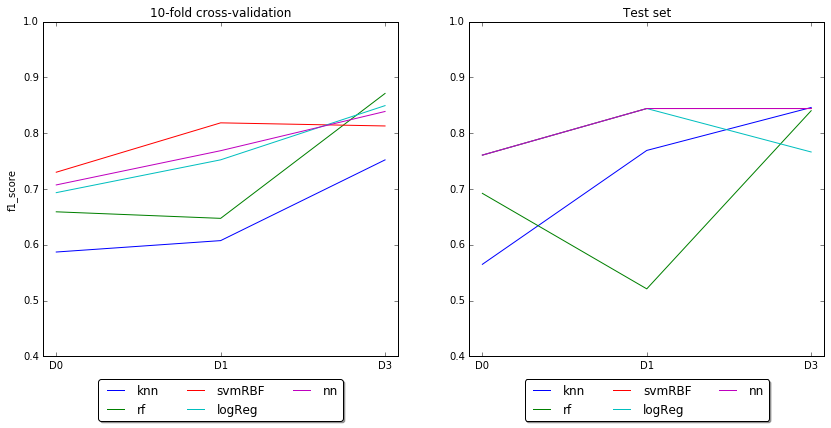}
  \caption{Performance results reached in cross-validation and test by the best pipelines at the different time-points..}
  \label{fig:topClsPerf}
  \end{figure}

We also analyzed what was the contribution in classification performance of the different techniques configured in the pipelines (i.e. sampling strategy, feature selection, learning metric and classification algorithm). In particular, the average in the performance (i.e. f1-score) of all pipelines using the sampling strategy was (0.59+/-0.07,0.61+/-0.09,0.75+/-0.09) for each dataset. In contrast, not using any sampling strategy was (0.58+/-0.07,0.62+/-0.08,0.75+/-0.08). Concerning the average performance reached among the pipelines configured with the best feature selection methods they scored (0.59+/-0.03,0.63+/-0.07,0.77+/-0.08). In contrast, the average in performance reached by the pipelines without using any feature selection was (0.66+/-0.07,0.68+/-0.11,0.77+/-0.09) respectively. Focusing in the evaluation metric with which the pipelines were configured, the pipelines with f1-weighted metric achieved an average in the performance of (0.58+/-0.07,0.62+/-0.09,0.76+/-0.09) while the pipelines configured with precision-weighted obtained an average performance of (0.59+/-0.07,0.61+/-0.09,0.75+/-0.09). Regarding the type of classification algorithm (figure \ref{fig:topClsPerf}), the best pipelines reported performances in cross-validation between 0.59+/-0.21 (using k-NN) and 0.73+/-0.18 (using SVM) of f1-score in D0. In D1 the best pipelines reported performances between 0.61+/-0.12 (using K-NN) and 0.82+/-0.06 (with SVM). In D3 the best pipelines reported performances between 0.75+/-0.07 (using k-NN) and 0.87+/-0.15 (with RF). Table \ref{tab:diffPerfConfig} summarizes these differences of performance among the configured pipelines.

\begin{table}[ht!]
\caption{Performance difference of f1 cross-validation along the different datasets achieved  among pipelines configured with different techniques.}
\label{tab:diffPerfConfig}
\begin{center}
\begin{tabular}{ll|ll|ll|ll}
\hline
{} & {} & D0 & {} &  D1 & {} & D3  & {} \\
\hline
Methods & Comparisons & Avg &  Max & Avg & Max & Avg & Max \\
\hline
Sampling & Smote vs None &   0.01 &      -0.02 &  -0.01 & 0.04 & 0.0 & 0.02 \\
Feature Selection & Best vs  None &  -0.07 &      -0.07 &  -0.05 & -0.13 & 0.0 & 0.02 \\
Metrics & f1 vs prec &   0.01 &      0.02 &  0.01 & 0.04 & 0.01 & -0.01 \\
Classifier Algorithm & Best vs Worst &   0.09 &      0.14 &  0.14 & 0.21 & 0.19 & 0.20 \\
\end{tabular}
\end{center}
\end{table}

Regarding the comparison of performance between descriptive and non-descriptive pipelines, the best descriptive pipelines obtained f1-scores of 0.69 +/-0.15, 0.75 +/- 0.15 and 0.87 +/- 0.15 in cross-validation and 0.76, 0.84, 0.84 in test set, while the best non-descriptive pipelines  obtained scores of 0.73 +/-0.18, 0.82 +/- 0.06 and 0.84 +/- 0.08 in cross-validation and 0.76, 0.84, 0.84 in test set. Further details about these pipelines can be found in table \ref{tab:PipeDescNon}.

\begin{table}[ht!]
\caption{Best descriptive and non-descriptive pipelines by dataset.}
\label{tab:PipeDescNon}
\begin{center}
\begin{tabular}{llllllllllrrr}
\hline
ds &     sm &        fs & metric &  cls &                         params \\
\hline
D0 &   none &      none &   prec &  SVM &          [0.001, balanced, 30] \\
D0 &  Smote &      none &   prec &   LR &              [None, 15, 4, l2] \\
D1 &  Smote &      none &     f1 &  SVM &           [0.001, None, 4, 15] \\
D1 &   none &      none &     f1 &   LR &                  [None, 5, l2] \\
D3 &  Smote &  lasso\_fs &   prec &   RF &  [1, 250, gini, 4, None, None] \\
D3 &   none &      none &     f1 &   LR &                [None, 0.5, l1] \\
\hline
\end{tabular}
\end{center}
\end{table}

\begin{figure}[h!]
   \includegraphics[width=1.0\textwidth]{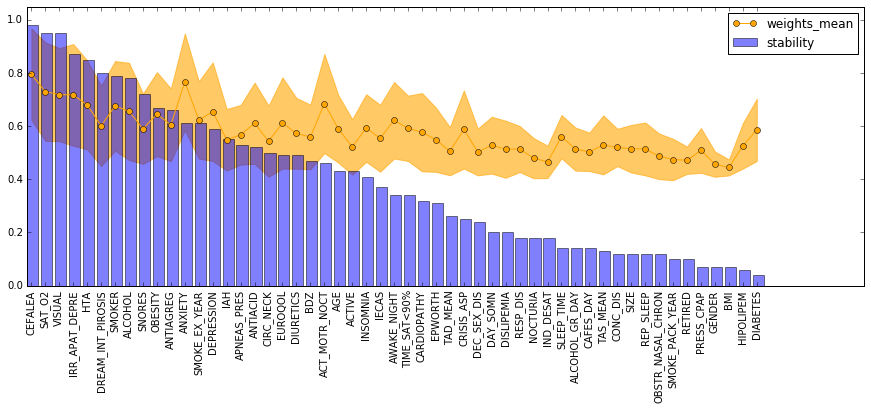}
  \caption{Stability scores and feature weights of the best pipeline for dataset D0.}
  \label{fig:d0Feats}
      \end{figure}

\begin{figure}[h!]
   \includegraphics[width=1.0\textwidth]{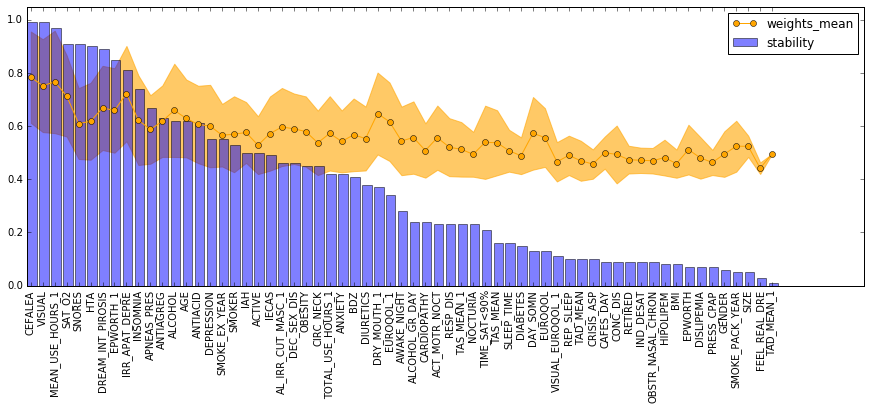}
  \caption{Stability scores and feature weights of the best pipeline for dataset D1.}
  \label{fig:d1Feats}
      \end{figure}

\begin{figure}[h!]
     \includegraphics[width=1.0\textwidth]{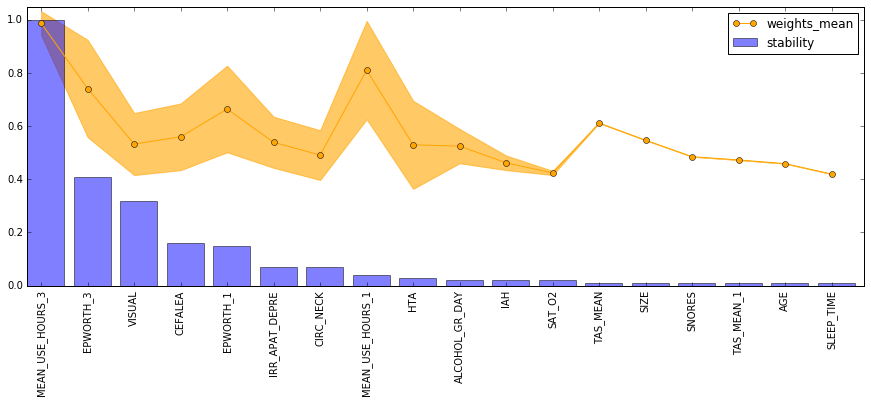}
  \caption{Stability scores and feature weights of the best pipeline for dataset D3.}
  \label{fig:d3Feats}
      \end{figure}

To complete the analysis we extracted the most important features used by the best descriptive pipelines. In total (25/28/20) features were reported for each dataset after setting up a minimum threshold of 0.4 in the feature weights. Top-10 features of D0 and D1 reported stability scores above 0.6. In D3 only 2 features were above 0.4 of stability score. Figures (\ref{fig:d0Feats}, \ref{fig:d1Feats}, and \ref{fig:d3Feats}) provides a visual ranking of the features of each of the dataset ordered by the stability score. 

\section{Discussion}\label{sec:discussion}

This study is the first attempt to predict compliance with the CPAP therapy in patients with OSA at different points of the treatment by building classifiers.  A good CPAP compliance has been demonstrated to reduce cardiovascular risk and symptoms in patients with sleep apnea. The identification of a specific group of patients with a predicted poor compliance could allow focusing resources and capitals on this specific group for improving its compliance.

Results have shown that the D0 dataset, collected before the start of treatment, is the most complex to learn compared to D1 and D3, collected at months 1 and 3 of the patient's therapy. Nevertheless, a considerable average f1-score of 0.73 +/- 0.18 was achieved in cross-validation and 0.76 in the test set. As shown in table \ref{tab:diffBestPipes}, an important performance increase of 0.09 (p = 0.12) was reached in D1 with respect to D0. In D3, we get the best classification performance with a significant increase of 0.14 in f1-score (p = 0.024) with respect to D0. This same trend occurred in the test set where the best pipelines of D1 and D3 reported performances of 0.18 above the achieved in D0. The difference in f1-score between D3 and D1 did not prove to be significant (p = 0.30). These results seem to confirm that follow-up measurements help to increase baseline prediction performance. Indeed the closer to the CPAP compliance cut-off we are the more confident is the classifier (i.e. performance in D0 is minor than performance in D1 and minor than D3). In addition to that, patterns of CPAP adherence appear early, in our case at 1 month, since it is when the greatest performance increase is achieved. This same finding was confirmed by other studies \cite{bib:budhiraja07} \cite{bib:weaver1997night}. 

During the evaluation step, we realized the use of sampling, feature selection or a particular learning metric were not as substantial as expected in any of the datasets (table \ref{tab:diffPerfConfig}). To be more specific, the maximum performance increase, regardless of the method used, was between 0.02 and 0.04 of f1-score in cross-validation. Probably this confined contribution was due to the initial preprocessing and the fact that the data were not severely unbalanced. Indeed, in D0 and D1 the use of feature selection compromised the performances with a maximum decrease of 0.13. In contrast, important increments of performance in all datasets were produced depending on the classification algorithm used (i.e.  0.14, 0.21 and 0.20). In D0 and D1 best pipelines were using an SVM. This result was not surprising because this algorithm has been already reported suitable for problems with few samples and with a high number of features \cite{bib:scholkopf2002learning} being able to build complex non-linear decision boundaries. In D3 the best pipeline was configured with an RF although the one with SVM also provided a high score. RF algorithm is also suitable for difficult problems and especially indicated for handling categorical features \cite{bib:ho95}. The other non-descriptive classifier (i.e. NN) reported competent performances in all three datasets, especially in D3, where it exceeded the results reported by the best pipeline configured with an SVM. However, the K-NN algorithm reported the worst scores on the three data sets. This is partly because it does not usually work well with a large number of features \cite{bib:weber1998quantitative}.

Focusing on the differences of performance between the best descriptive and non-descriptive pipelines, those were always below 0.1 and not significant (p=0.14) for D0 but significant in D1 (p=0.02). In contrast,  the best performance in D3 was achieved through a descriptive classifier although in cross-validation the difference in performance with the best non-descriptive pipeline (0.02+/-0.16 of f1-score) proved to be non-significant (p=0.69).

Regarding the relevant factors related to the CPAP compliance prediction, four baseline features were found common in each of the best pipelines for the different datasets collected at time-points (T0, T1, and T3). Those were headaches, psychological symptoms (i.e. irritability, apathy, and depression), arterial hypertension and the visual analog scale (as part of EuroQol questionnaire). From these four features, the headache was the most stable feature (i.e. with the highest stability score) at T0 and T1. In the baseline, all these characteristics were found significant with respect to the CPAP compliance but the latter (p=0.079). In particular, compliant patients were more likely to not having headache (85\%, 23 out of 27) nor psychological symptoms (67\%, 18 out of 27), having arterial hypertension (74\%, 20 out of 27) and worst visual analog scale score (9.15+/-1.02 on mean difference with respect to non-compliant patients). To our knowledge, these features together have not previously been reported as relevant to predict patient compliance with CPAP therapy at either month 0, 1 and 3. In the literature, having morning headache was also found significant in a randomized control trial of OSA patients \cite{bib:kristiansen2012sleep}. In contrast, psychological factors did not show prediction capability in \cite{bib:stepnowsky02}\cite{bib:aloia05}\cite{bib:mostofsky2014handbook} but how patients were challenging difficult situations (active versus passive) \cite{bib:stepnowsky2002psychologic}. In \cite{bib:martinez2013effect}, authors evidenced the positive effect of CPAP treatment on blood pressure in patients with resistant hypertension. The visual analog scale, used as a generic method for measuring the quality of life, was reported useful to track treatment-induced changes in \cite{bib:schmidlin2010utility}\cite{bib:chakravorty2002health}.

Different studies \cite{bib:tamanna2016effect}\cite{bib:li2014self} \cite{bib:campos2005mortality} have shown an improvement in snoring, gastroesophageal reflux and oxygen saturation with CPAP treatment. In our sample, only oxygen saturation ($p<0.001$) predicted good compliance with CPAP. However, these were found among the characteristics with the highest stability scores for the best pipelines of month-0 and month-1.

Two of the features collected at months 1 and 3 (i.e. average hours of nightly CPAP use and Epworth) were found among the most important predictive features in these time-points. These features were significant regarding CPAP compliance. Interestingly, from months 1 to 3 the average of nightly hours of use for compliant users increased (from 5.9+/-1.51 to 6.17+/-1.29) while in non-compliant users decreased (from 4.4+/-1.75 to 3.56+/-1.76). In contrast, the average of Epworth for compliant users decreased from 5.48+/-3.63 to 4.64+/-3.07, while for non-compliant increased from 7.33+/-3.7 to 8.46+/-4.16). Early measurements of the average hours of nightly CPAP use were already reported as predictive of CPAP compliance in \cite{bib:chai2013predictors}\cite{bib:ghosh2013identifying}. Epworth was also reported as a relevant predictor of compliance  in \cite{bib:budhiraja07}\cite{bib:popescu2001continuous}\cite{bib:weaver2007relationship}\cite{bib:ghosh2013identifying}.

The limitations of this study come from two sides. First, although a common cut-off was selected for the definition of CPAP compliance, changes in this threshold might cause different performances as well as variations in the rank of the feature importance reported in this work, thus further explorations are required in this regard. Second, even the positive scores obtained by the predictive models at the different time-points, the reduced number of individuals in the sample produces appreciable variations in their performance. Therefore collecting more data could help to validate the presented findings. 

\section{Conclusions}
To the best of our knowledge, this article is the first attempt to analyze and compare the compliance with the CPAP therapy of patients with OSA at different points of the treatment by building classifiers. Three time-points were established to perform the analysis (i.e. before the treatment starts, after one month and at the month 3). To build and evaluate the classifiers a flexible framework was designed relying on machine learning pipelines. High performances were reached yet after one month of treatment, being the third month when significant differences in performances were achieved with respect to the baseline. Four baseline variables were reported relevant for the prediction of compliance with CPAP at each time-point. Two characteristics more, collected during the follow-up, were also highlighted for the prediction of compliance at months 1 and 3. Further tasks are devised to extend the present study, including the collection of new patients and exploring other CPAP compliance cut-offs, in order to validate the findings and reported performances. This work has intended to take a step forward towards the creation of new tools to allow early and accurate detection of patients struggling to follow the CPAP treatment and thus enable personalized patient interventions that would lead to improving their quality of life.

\section*{Declarations}
\subsection*{Ethics (and consent to participate)}
The study was approved by the hospital’s ethics committee (26/06/2014, num.10/2014)  All recruited patients signed an informed consent form.

\subsection*{Consent to Publish}
Not applicable.

\subsection*{Competing interests}
The authors declare that they have no competing interests.

\subsection*{Author's contributions}
XRP performed the analysis of the datasets and worked on the model definition and creation. CT gave clinical support in all the phases of the model definition and creation. AS contributed to data collection, labeling and pre-processing. MST and FB supervised the study from the clinical perspective. EV scientifically supervised the study.

\subsection*{Availability of Data and Materials}
All relevant data are in the manuscript

\section*{Acknowledgements}
This work is part of the myOSA project (RTC-2014-3138-1), funded by the Spanish Ministry of Economy and Competitiveness (Ministerio de Econom\'{i}a y Competitividad) under the framework ``Retos-Colaboraci\'{o}n'', State Scientific and Technical Research and Innovation Plan 2013-2016.

\bibliographystyle{bmc-mathphys}
\bibliography{references}   

\newpage
\section*{Figures}

\end{document}